\documentclass[letterpaper]{article} 
\usepackage{aaai25}  
\usepackage{times}  
\usepackage{helvet}  
\usepackage{courier}  
\usepackage[hyphens]{url}  
\usepackage{graphicx} 
\usepackage{natbib}  
\usepackage{caption} 
\usepackage{algorithm}
\usepackage{algorithmic}

\usepackage{newfloat}
\usepackage{listings}
\DeclareCaptionStyle{ruled}{labelfont=normalfont,labelsep=colon,strut=off} 
\floatstyle{ruled}
\newfloat{listing}{tb}{lst}{}
\floatname{listing}{Listing}
\usepackage{pifont}
\usepackage{amsmath}

\newcommand{\cmark}{\ding{51}}%
\newcommand{\xmark}{\ding{55}}%

\title{Multi-Agent Reinforcement Learning with Long-Term Performance Objectives for Service Workforce Optimization}

\author{
  Kareem Eissa, Rayal Prasad, Sarith Mohan, Ankur Kapoor, Dorin Comaniciu, Vivek Singh
}
\affiliations{
  Siemens Healthineers\\
  Digital Technology and Innovation \\
  Princeton, NJ, 08540, USA \\
  \{kareem.abdelrahman,rayal.prasad,sarith.mohan\}@siemens-healthineers.com \\
  \{ankur.kapoor,dorin.comaniciu,vivek-singh\}@siemens-healthineers.com \\
}

\usepackage{bibentry}

\begin{document}

\maketitle

\begin{abstract}
Workforce optimization plays a crucial role in efficient organizational operations where decision-making may span several different administrative and time scales. For instance, dispatching personnel to immediate service requests while managing talent acquisition with various expertise sets up a highly dynamic optimization problem. Existing work focuses on specific sub-problems such as resource allocation and facility location, which are solved with heuristics like local-search and, more recently, deep reinforcement learning. However, these may not accurately represent real-world scenarios where such sub-problems are not fully independent. Our aim is to fill this gap by creating a simulator that models a unified workforce optimization problem. Specifically, we designed a modular simulator to support the development of reinforcement learning methods for integrated workforce optimization problems. We focus on three interdependent aspects: personnel dispatch, workforce management, and personnel positioning. The simulator provides configurable parameterizations to help explore dynamic scenarios with varying levels of stochasticity and non-stationarity. To facilitate benchmarking and ablation studies, we also include heuristic and RL baselines for the above mentioned aspects.
\end{abstract}

%

\section{Introduction}
Uncovering strategies to achieve operational efficiency and fairness in service delivery or management has been a topic of longstanding research, with ubiquitous applications ranging from supply chain logistics to healthcare delivery [\cite{jaillet22}]. The past several years have seen the development of digital twins that model operational infrastructure for their service operations, and use heuristics or other developed methods to discover potential workflow improvements. Such efforts span multiple domains such as service delivery with offerings from ServiceNow, ServiceMax, Oracle, supply chain from IBM and LightSource, industry from Siemens and GE, and hospital operations from Palantir, LeanTaaS among others. In most cases, however, the focus is often on the data mapping and providing interfaces for user driven analysis. While this leads to improvements in desired KPIs (Key Performance Indicators), the impact is often limited since these decisions are made in silos to keep data complexity manageable. For instance, the task responding to service requests is often analyzed separately from having sufficient service personnel with desired experience at desired locations. In reality, however, these are strongly coupled with ensuring service quality over a period. Keeping this in mind, we believe that having an AI gym-like environment for the management of service resources that incorporates dynamic and non-stationary aspects of the real world would enable researchers to develop AI solutions that bring greater positive impact.


Workforce management scenarios can vary significantly in scale, and thus having a modular simulation environment is key to accommodating various scenarios. Furthermore, real-world conditions of workforce management are highly dynamic and it is important for the environment to support a high degree of fidelity. Finally, it is essential that the environment supports the development of reinforcement learning (RL) for high-dimensional optimization problems - it has been shown that RL achieves state-of-the-art performance on long-horizon problems in large-scale dynamic environments that have multiple objectives [\cite{lau2022}], and it aligns well with the workforce optimization formulation that would typically entail multiple objectives.


In this work, we aim to bridge the gap between RL and workforce management. We introduce a parameterized modular environment that can be configured to simulate a range of scenarios while also providing granular control over stochasticity, non-stationarity, and scale of the simulation. Additionally, we provide baseline heuristics for each task in the environment to facilitate ablated performance comparison of different subsets of tasks measured by a set of workforce-related metrics. Finally, we benchmark RL approaches and show promising results that perform favorably against the heuristics but with room for improvement in future research. We also compare performance to methods developed in isolation on a subset of tasks to highlight the advantages of solving workforce optimization as a joint problem in the integrated setup. 

\begin{table*}[h!]
\centering
\caption{Comparison of various simulation environments for operations management.}
\begin{tabular}{|l|c|c|c|c|c|}
\hline
Environment  & Supports  & Short-Term & Supports & Collective  & Long-Term  \\
& Single-Agent & Decisions & Multi-Agent & Decisions & Decisions \\
\hline
\hline
OR-gym & \cmark & \cmark & \xmark & \xmark & \xmark \\
ORL [AAAI'20] & \cmark & \cmark & \xmark & \xmark & \xmark \\
ORSuite [SIGMETRICS'22]& \cmark & \cmark & \xmark & \xmark & \xmark \\
MARO [AAMAS'19] & \cmark & \cmark & \cmark & \xmark & \xmark \\
OFCOURSE [NeurIPS'23] & \cmark & \cmark & \cmark & \cmark & \xmark \\
Our work & \cmark & \cmark & \cmark & \cmark & \cmark \\
\hline
\end{tabular}
\label{table:related_work}
\end{table*}

\section{Related Work}
Managing service workforce to deliver reliable and timely service to a large number of facilities involves addressing several different aspects, many of which has been researched albeit often in silos. A key aspect is to optimize the assignment of personnel to service requests to achieve a high throughput, which can be formulated as an assignment problem [\cite{pathan21}]. Another aspect is that of managing workforce personnel (such as making hiring decisions) which addresses the trade-off between workforce utilization and cost, and can be viewed as the well-studied bin packing problem while minimizing the number of bins. Since facilities are often distributed over a large geographical area, positioning of new personnel to reduce travel while ensuring a low aggregate workforce cost is similar to the facility location problem [\cite{guo23swapbasedDR}]. Another aspect of managing workforce training and qualifications is tied to the RL literature on churn analysis [\cite{panjasuchat20}]. While such formulations help address one aspect of workforce optimization, they often result in sub-optimal performance metrics in others.

Recent literature on resource optimization has seen a growth in development of simulation environments that model the interdependence between various aspects and enable researchers to develop and evaluate RL algorithms for operations research (OR). [\cite{orgym}]'s OR-Gym provide environments for the knapsack, bin packing, supply chain, and asset allocation (assignment) problems. OR-Suite by [\cite{orsuite}] aims to facilitate RL research on OR problems such as ambulance routing, fair resource allocation for food-banks, vaccine allocation in pandemics, and ridesharing. [\cite{maro}] introduced MARO (Multi Agent Resource Optimization) benchmark for RL as a Service for real-world resource optimization and includes different usecases such as container inventory management in logistics and bike repositioning in transportation. Recent work by [\cite{ofcourse}] recognized the limitations of these frameworks on still being limited in terms of covering the holistic nature of these scenarios, and presented OFCOURSE, a holistic environment to model the entire supply chain, with emphasis on the order fulfillment problem. However, unlike order fulfillment, a holistic workforce optimization environment needs to model both high frequency decisions on dispatch as well as low frequency decision on workforce changes and their long term interdependence. While the existing frameworks are not well-suited to support holistic workforce optimization, there have been valuable in progressing RL research. For instance, the MARO benchmark was leveraged by [\cite{sinclair23}] for hindsight learning that uses optimization solver on hindsight data as an oracle to train an RL that outperforms state-of-the-art methods. Our aim with providing a simulator for workforce optimization is to further facilitate such research to address the problems more holistically. A feature-wise comparison with the other simulation environments is shown in Table \ref{table:related_work}.

\section{Simulator design}

\subsection{Motivating Example: Service workforce management}
Consider an organization that provides a service to maximize the machine uptime of equipment such as radiology machines at hospitals or energy management equipment at hotels. Such organizations have a workforce of service engineers trained and qualified to operate such machinery, and are at different levels of experience and certification [\cite{wo18stepguide,decisionbrain19,oracle21}]. Under ideal settings where equipment is perfectly reliable and has no downtime, these engineers are waiting idle at their base office location. When an equipment is under distress or malfunctions, a qualified engineer that is available is dispatched to conduct repairs on-site. While dispatching the engineer, it is important to factor in their experience level in addition to their travel time to the facility location. Engineers with more experience often would address the issue in one (or a few) visits compared to novice ones who may need multiple visits to have time to consult with experts, retrieve the right materials, rework their previously failed service, etc.. Needless to say, the availability of an engineer, who is also strategically at a favorable location, is also tied to the number of service personnel in the organization. Hence, dealing with task of ensuring high uptime for equipment involves simultaneous optimization of the dispatch strategy, staff count, their expertise and location, among other factors. Furthermore the mean time between such malfunctions across all equipment in the field can not be reliably estimated and likely follows a non-stationary distribution, which further complicates the optimization process.

\begin{figure*}[ht]
    \centering
    \includegraphics[width=\linewidth]{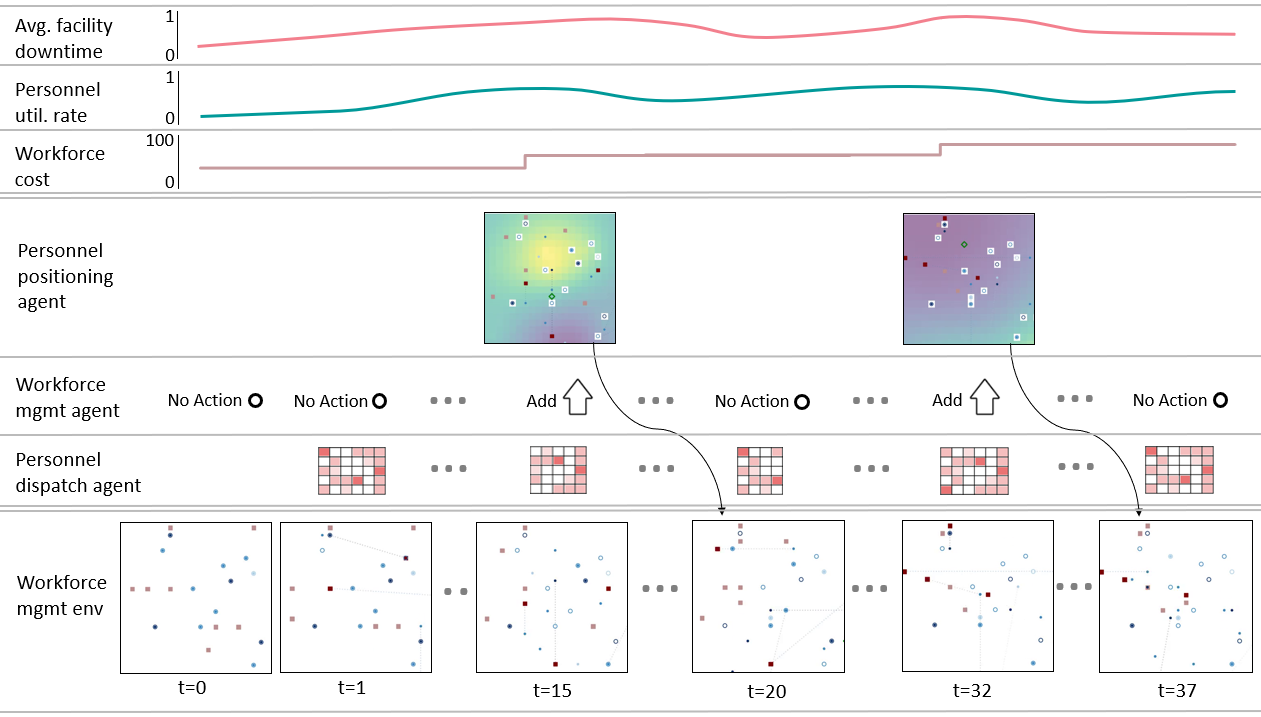}
    \caption{A simulated run of the workforce management environment. Each time-step has a corresponding environment state showing the workforce ($N_t$ circles), facilities ($M_t$ squares) and assignments (lines). Notice that personnel dispatch agent take frequent actions ($N_t$ x $M_t$ assignments) to dispatch personnel to appropriate facilities. The workforce management agent, on the other hand, often takes no action, and when it does, the impact of the action is observed several time steps later.}
    \label{fig:overview}
\end{figure*}


\subsection{Problem formulation}

\subsubsection{Simulation Entities}

\noindent \textbf{Workforce} Service personnel that execute service requests at facilities, represented by engineers in the above example. They have an experience level that affects their efficiency, while also altering their workforce cost. Each personnel has a fixed home office location that is determined upon hiring. They return to their home office location after each dispatch to a facility.

\noindent \textbf{Facility} Entities with fixed locations that require servicing by the workforce, represented by equipment in the above example. New facilities can become active and active ones can go offline throughout a simulation.

\subsubsection{Metrics}
We describe the 3 metrics used as key performance indicators (KPIs) that assess the trade-offs between facility downtime, personnel costs and balanced workload.

\paragraph{Workforce cost ($\operatorname{WC}$)} The cost associated with maintaining a certain number of personnel on the workforce. The organization would aim to minimize this cost while balancing the other objectives. We compute this metric as the size of the workforce in log scale.
\begin{equation}
    \operatorname{WC} = log(|\operatorname{Personnel}|)
\end{equation}

\paragraph{Personnel utilization rate ($\operatorname{PUR}$)} This metric is associated with workload distribution fairness and utilization of working hours. An employee should not be overworked and utilized around some threshold e.g., 75\%. We modify this metric using a rolling window to accommodate long-horizon episodes - in such cases, the full-episode metric would saturate as the horizon increases since any actions would have diminishing influences. We measure this using the formula below where $\operatorname{Work}$ is the number active working (i.e., travel or service) time-steps within the rolling window and $T_{\operatorname{PUR}}$ is the size of the rolling window for this metric.
\begin{equation}
    \operatorname{PUR} = \frac{\operatorname{\operatorname{Work}}}{T_{\operatorname{PUR}}}
\end{equation}

\paragraph{Average facility downtime ($\operatorname{AFD}$)} This is a facility metric that provides a measure of downtime and is analogous to meeting demand and customer satisfaction. Similar to the \textit{personnel utilization rate}, we consider a rolling window to balance short and long term effects. In the formula below used to measure this metric, $\operatorname{Downtime}$ is the active wall-clock time while the service request remained unfulfilled within the rolling window and $T_{\operatorname{AFD}}$ is the size of the rolling window for this metric.
\begin{equation}
    \operatorname{AFD} = \frac{\operatorname{Downtime}}{T_{\operatorname{AFD}}}
\end{equation}

\subsubsection{Action space}

We model 3 aspects of control, each of which is provided with a heuristic baseline that can be used to train models for other aspects only. Additionally, these heuristic baselines can be configured to simulate deterministic behavior which would help in studying the steady-state characteristics of the simulation.

\noindent \textbf{Personnel dispatch} Assigning workforce personnel to facility service requests. This can be repair engineers responding to machine failures or even radiology centers responding to patient appointments. The decision-making for this aspect revolves around a many-to-many assignment.

\begin{figure*}[ht]
    \centering
    \includegraphics[width=0.85\linewidth]{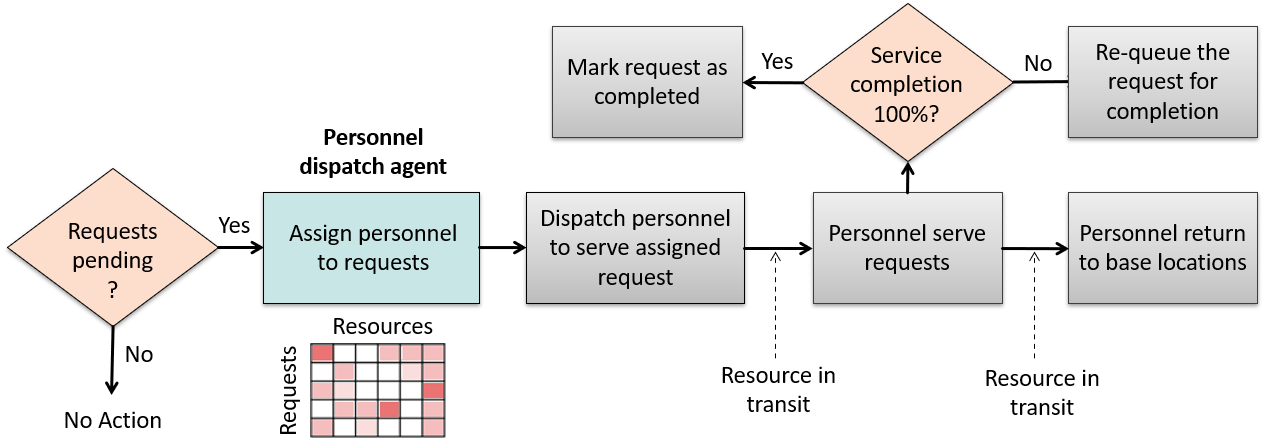}
    \caption{Process flow for personnel dispatch to service requests generated by facilities.}
    \label{fig:dispatch_agent}
\end{figure*}

\noindent \textbf{Workforce management}
Changing the size of the workforce - the decision for this aspect is only concerned with \textit{when} to change the size of the workforce, and whether to increase or decrease at specific levels of expertise.

\noindent \textbf{Personnel positioning}
Decision-making about \textit{where} to make changes to the workforce. This aspect is only applicable when a workforce management action is triggered making them strongly coupled.

\subsection{Discrete-Event Modeling}

There is a long-standing history of using Discrete-Event Simulation (DES) to model various business and industrial processes [\cite{mockett08,amrouni21,zinoviev2024}]. DES allows for control over the realism of the simulation through the complexity of the modeled processes. We leverage the DES formalism for workforce optimization.
We abstract some aspects of the motivating example and aim to design a general simulator for different workforce optimization scenarios. However, we keep some key components to help capture basic aspects of the problem. 

DES operates by processing triggering events and updating the state of the simulation accordingly. For our simulator, the primary trigger is for a facility to request service e.g., machine in need of maintenance. Figure \ref{fig:dispatch_agent} illustrates the control flow for this trigger. We model these service events  by a Poisson process with a controllable inter-arrival rate. The DES then receives a personnel dispatch action (whether for a learned or heuristic policy) which it updates the state of the simulation while handling action validation and race conditions e.g., cannot dispatch the same person to multiple facilities simultaneously. At each time-step, the DES simulates personnel travel between facility locations and their home office locations. 

\begin{figure*}[ht]
    \centering
    \includegraphics[width=0.85\linewidth]{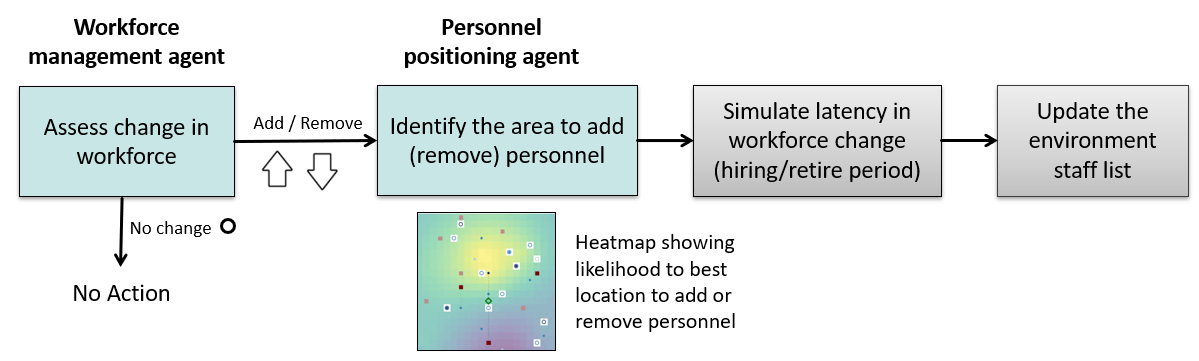}
    \caption{Process flow for workforce management and personnel positioning agents to add or remove personnel. Note that when a new personnel needs to be added, the personnel positioning agent identifies the area where to hire, and when personnel count needs to be reduced, the personnel positioning agent selects the personnel by identifying the area where reduction would help.}
    \label{fig:staff_agents}
\end{figure*}

The service requests and personnel dispatch operate at a high frequency. At a lower frequency, the DES also receives workforce management actions to alter the size of the workforce e.g., hire new personnel. Figure \ref{fig:staff_agents} illustrates the control flow for these actions. Alongside this action is a positioning action which specifies where to alter the workforce. These actions are not triggered but initiated by the policy. Workforce management and positioning actions have a delay before they are executed to simulate on/off-boarding process.

Another trigger in the environment is a change in facilities to simulate acquiring or losing clients. We also model this by a Poisson process controlling the frequency of facilities entering or exiting the state of the simulation.

\subsubsection{Environment parameterization}
DES provides a powerful tool to investigate a system under different conditions and what-if scenarios. To facilitate such study of the workforce optimization problem, we design our simulator with parameters to control different aspects of the simulated environment.

Our simulator allows running scenarios with varying number of facilities and personnel. We model service requests as a Poisson process which parameterizes the inter-arrival rate of requests. For facilities we model entering and leaving the simulation, also parameterized by a Poisson process. The size of the rolling window controls the effective horizon of the metrics, allowing a trade-off between myopic and saturated behavior. Finally, we parameterize the environment grid size to explore the aspects of spatial scaling.

\section{Experiment design}
To showcase the functionality of the simulator and the effectiveness of RL in solving the joint optimization objective, we describe two baseline approaches (heuristic-based and RL) in this section along with some experimental results.

\subsection{Baselines: Heuristics}
\noindent \textbf{Personnel dispatch} 
We design a greedy distance-based heuristic where service personnel are assigned to their nearest facility with an open service request. We implement a stochastic variant with weighted sampling inversely proportional to the distance. This variant is useful to study the steady-state of dispatch, especially in a deterministic setting.

\noindent \textbf{Workforce management}
We design a greedy threshold-based heuristic that triggers an increase in the number of personnel if the resources are over-utilized and a decrease in the number of personnel if they are under-utilized. Similar to the dispatch heuristic, this has a stochastic variant that implements an $\epsilon$-greedy sampling between the threshold-based decision and a random decision (increase, decrease, do-nothing). With multiple levels of personnel experience, this heuristic chooses the level of experience randomly while being  biased towards the middle level of experience.

\noindent \textbf{Personnel positioning}
We design a greedy spatial averaging heuristic corresponding to workforce personnel increase or decrease. Each facility computes a service demand metric modeled on it's history of service requests and downtime. We use this service demand as the weighing metric i.e., map of service quality, to position newly hired personnel. Similar to the previous heuristics, this baseline has a stochastic variant that samples around a smoothed Gaussian instead of a deterministic singular point which can be seen in the heatmap in Figure \ref{fig:staff_agents}.

\begin{figure*}[ht]
    \centering
    \includegraphics[width=0.9\linewidth]{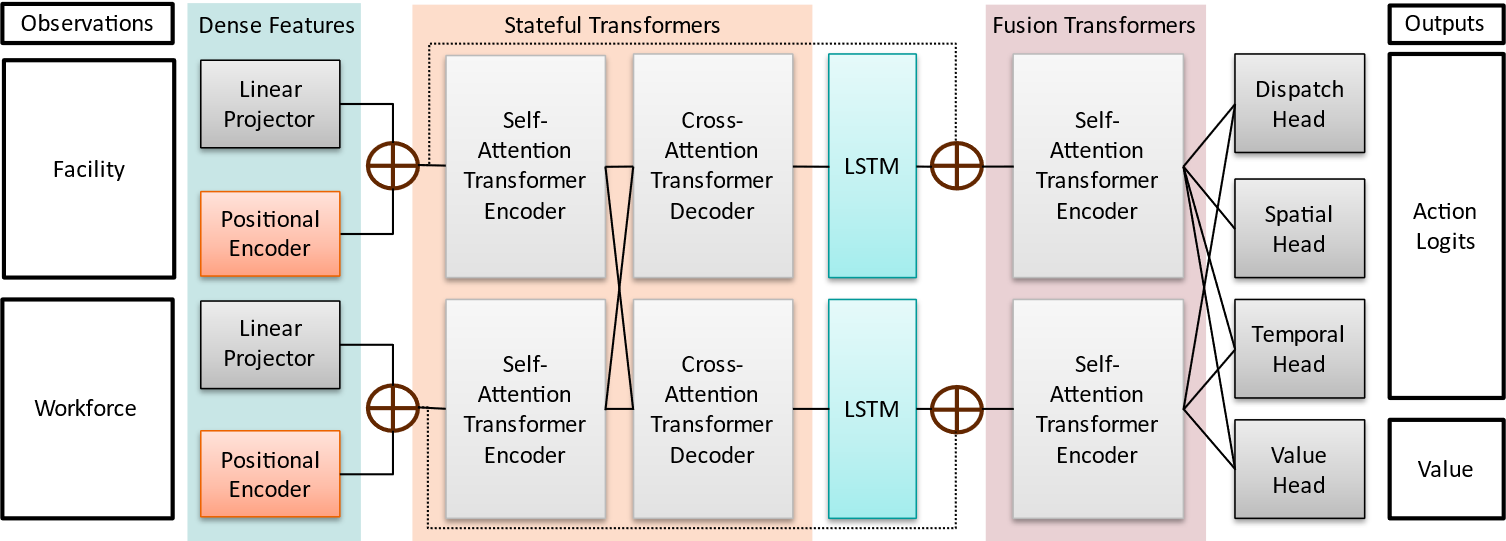}
    \caption{Network Architecture. Note that the spatial output head does not leverage the workforce features since it only involves a voting mechanism among the facility locations.}
    \label{fig:arch}
\end{figure*}

\subsection{Baselines: Reinforcement Learning}

\noindent \textbf{Observations} The observation space comprises of features from both the facilities and the workforce. For the facilities, the observation space is a matrix of size $M$x7 where $M$ is the number of facilities. Each facility feature vector consists of (1-2) the facility's $XY$ location, (3-5) whether it is operational, requests service, or has a personnel assigned, (6) assigned personnel identifier, and (7) rolling-window of downtime. Similarly for the workforce, the observation space is an $N$x9 matrix where $N$ is the number of personnel. The feature vector for each personnel consists of (1-2) home office $XY$ location, (3-4) current travel $XY$ location, (5) level of expertise, (6) whether idle or assigned, (7) assigned facility identifier, (8) whether traveling to facility or to home office, and (9) rolling-window of utilization rate. 

\noindent \textbf{Rewards} To train multi-objective RL, we scalarize our metrics into a singular value using Nash Social Welfare [\cite{fan23welfare}] which has desirable properties such as monotonicity and Pareto optimality. This generalizes to other sets of metrics. For our experiments the reward is:

\begin{equation}
    R = \sqrt[3]{\operatorname{WC} * \operatorname{PUR} * \operatorname{AFD}}
\end{equation}

\noindent \textbf{Actions} There are three distinct actions associated with the RL model:

Personnel dispatch: Each facility requesting service receives one personnel from the available workforce, in addition to a no-op action.

Workforce management: Choice to increase or decrease personnel for each level of experience, in addition to a no-op action.

Personnel positioning: Given a \textit{workforce management} action, each facility provides a weight in the range [0,1] that is used to compute a location heat-map for either increasing or decreasing the workforce.

\noindent \textbf{Joint Model} The transformer-based network architecture that is used here is shown in Figure \ref{fig:arch}. Beginning with the facility and workforce observations, these inputs are individually passed through a multi-layer  projection and non-linear positional encoder to create dense embedded feature sets. Each row (facility or workforce) is processed independently.

These embedded feature sets are then passed through a series of transformers to ensure that the entire state at that time-step is attended to before being incorporated into a recurrent memory (Long short-term memory or LSTM). The stateful transformers consist of layers of facility-facility and workforce-workforce self-attention layers as well as facility-workforce and workforce-facility cross-attention layers.

After passing through the LSTM, the outputs are concatenated with the original embedded feature sets and then passed through another series of self-attention layers to fuse the historical features with the current time-step state [\cite{lim2019}]. The final layers consist of four multi-layer projection heads - three for actions related to the action space described in section 3.3.3, and a value output to set up a typical actor-critic mechanism. We train the model using PPO [\cite{ppo}].

\noindent \textbf{Dispatch-only Model} To study the effectiveness of modeling the integrated workforce optimization problem, we implement a variant of our model for the personnel dispatch problem. We use an identical architecture to the joint model expect that it has only the dispatch action head. We train this model to solve the personnel dispatching problem without considering any workforce changes. We modify the reward by removing the Workforce Cost from the geometric mean. We then deploy this trained model along the heuristic baselines for the Workforce Management and the Personnel Positioning to see how well it integrates.

\subsection{Results}

\subsubsection{Experiment settings} The simulation environment implements the service workflows, as referred to in the literature \cite{wo18stepguide}, \cite{decisionbrain19}, \cite{oracle21}.  Personnel have 3 levels of expertise (novice, mid-level, expert). The simulator implements the environment as service over a local geographical area as a uniform grid with a given number of facilities and service personnel.
Personnel expertise levels are initialized with a symmetric distribution favoring mid-level and equally weighing novice and expert with probabilities (0.25, 0.5, 0.25) respectively. Facilities enter and exit the system exogenously with an inter-arrival rate of 100 time-steps. Similarly, facilities require service with an inter-arrival rate of 40 time-steps. The base first visit repair rate is set to 0.7 which corresponds to (0.49, 0.7. 0.84) for novice, mid-level, and expert respectively. Finally, we set the rolling-window size for computing metrics to 50.

To validate the simulator implementation and association with the KPIs, we conducted several experiments to study the performance trends with varying grid/area size, number of personnel and facilities, and the performance trends appear to be consistent with the variations. Table \ref{table:trends} summarizes these trends, averaged over 50 runs, and appear to correlate with expected trends with increasing service personnel for a given number of facilities and vice versa.

\subsubsection{Quantitative comparison between policies}
To study the performance of our joint RL method compared to one trained on an isolated task and heuristic baselines, we consider the following environment parameterization - we set the grid size to 64x64 with a maximum of 50 facilities and 50 personnel initialized around 25 each. 

Table \ref{table:quantitative} shows a quantitative comparison between multiple RL algorithms (PPO \cite{ppo}, IMPALA \cite{impala}) as well as heuristics against the baseline where decisions are being made randomly. As expected, commonly used field heuristics outperform the baseline significantly. Notice that both PPO (on-policy) and IMPALA (off-policy) algorithms performed similarly but better than heuristics method, with PPO performing marginally better.

\begin{figure*}[ht]
    \centering
    \includegraphics[width=0.95\linewidth]{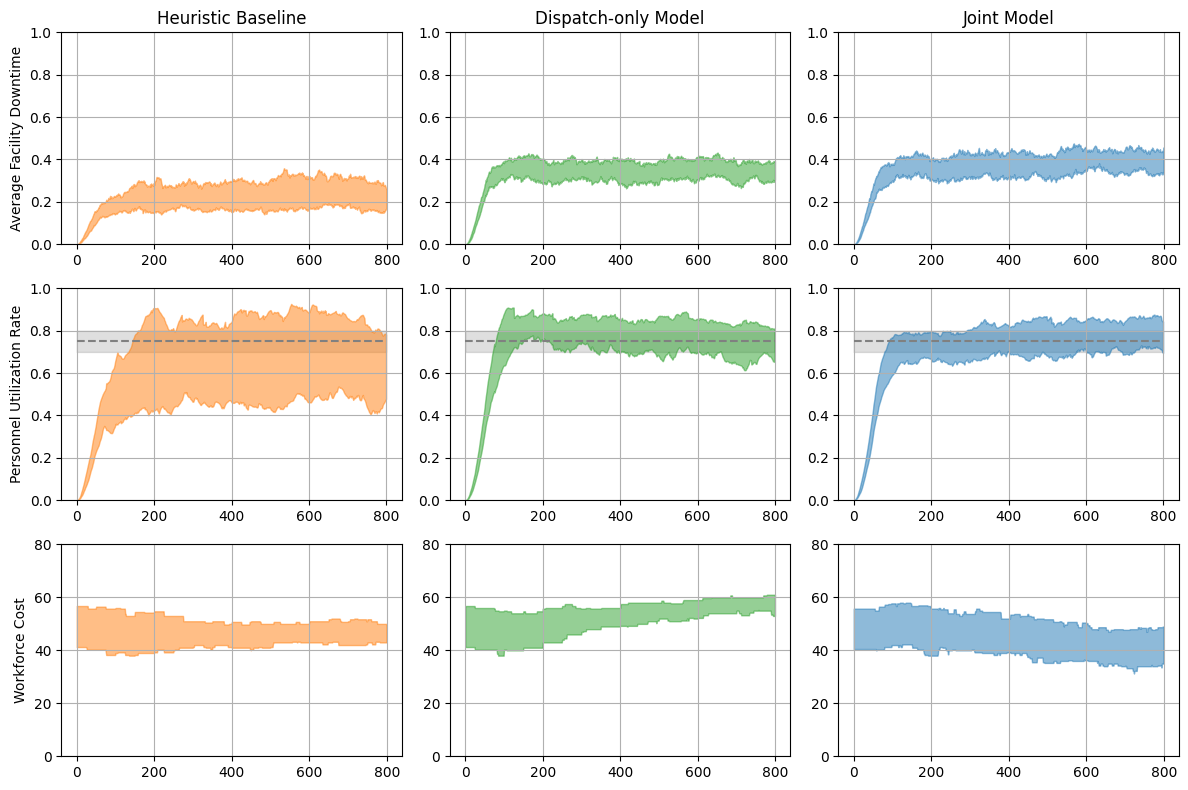}
    \caption{Results for scenario described in section 4.3. Each row reports on one of the metrics and each column reports on one of the methods. For the middle row, the gray band represents is ideal working conditions with the Personnel Utilization Rate near 0.75 with a margin of 0.05}
    \label{fig:results}
\end{figure*}

Figure \ref{fig:results} shows a compare the evolution of the KPIs over a period using methods 4, 5 and 6 from Table \ref{table:trends}. We report the $25^{th}$ to $75^{th}$ percentiles for the 3 metrics using the different methods over a horizon of 800 time-steps for 50 random episodes. All 3 methods perform well on the average facility downtime, where lower is better, with a marginal improvement for the heuristic baseline. As for the personnel utilization rate, where being closer to the gray band is better, both RL models perform better than the heuristic baseline with lower variance and more centered around the desired band. The workforce cost is lower for the joint RL model compared to the dispatch-only model.

As the dispatch-only model was trained in isolation from the integrated optimization problem, it is less adaptive to a dynamically changing workforce. We hypothesize that such a model tries to balance personnel utilization on average, without considering long-term personnel changes and how they affect utilization. 
On the other hand, a model trained to optimize the joint problem considers more diverse strategies such as hiring new personnel to balance utilization.

\begin{table*}[h!]
\centering
\caption{Nash social welfare (averaged over 50 runs) for varying grid size, number of service personnel (S) and facilities (F). 
Notice that for any given grid size, increasing service personnel with 25 facilities (first 3 columns) improves the KPI.
}
\begin{tabular}{|l|c|c|c|c|c|}
\hline
Grid size & 25F x 5S & 25F x 10S & 25F x 25S & 50F x 25S & 100F x 25S \\
\hline 
\hline 
32 x 32   & 5.53 & 3.16 & 2.96 & 2.91 & 6 \\
64 x 64   & 6.37 & 4.41 & 3.12 & 5.14 & 6.73 \\
128 x 128 & 6.7 & 6.05 & 3.65 & 6.37 & 7.17 \\
256 x 256 & 6.86 & 6.35 & 4.57 & 6.78 & 7.25 \\
\hline
\end{tabular}
\label{table:trends}
\end{table*}

\begin{table*}[h]
\centering
\caption{Quantitative comparison of algorithms for each agent. KPIs are averaged over 50 runs, and reported as "mean (25\% - 75\%)". 
}
\begin{tabular}{|c|l|l|l||c|c|c|c|}  
\hline
Method \# & Dispatch & Workforce & Personnel  & Workforce  & Personnel & Facility & Nash social \\
 & &  mgmt & positioning &  cost & util. rate & downtime & welfare \\
\hline
\hline
1 & Random & Random & Random & 75 (59-95) & 0.6 (0.4-0.8) & 0.3 (0.2-0.4) & 5.9 (2.4-11.1) \\
2 & Random & Heuristic & Heuristic & 57 (48-66) & 0.8 (0.5-0.9) & 0.3 (0.2-0.4) & 3.7 (1.8-5.7) \\
3 & Heuristic & Heuristic & Random & 45 (35-57) & 0.6 (0.3-1) & 0.2 (0.1-0.4) & 4.9 (2-7.5) \\
4 & Heuristic & Heuristic & Heuristic & 46 (42-50) & 0.6 (0.5-0.8) & 0.2 (0.2-0.3) & 4.1 (3-5.1) \\
\hline
5 & PPO & Heuristic & Heuristic & 57 (51-61) & 0.7 (0.7-0.8) & 0.3 (0.3-0.4) & 3.5 (2.4-4.4) \\
6 & PPO & PPO & PPO & 45 (37-55) & 0.7 (0.7-0.8) & 0.4 (0.3-0.4) & \textbf{3.1 (2.2-3.7)} \\
7 & IMPALA & IMPALA & IMPALA & 49 (40-56) & 0.6 (0.6-0.7) & 0.4 (0.3-0.5) & 3.1 (2.5-3.5) \\
\hline
\end{tabular}
\label{table:quantitative}
\end{table*}

\section{Discussion}
While we have used the example of service workforce management to demonstrate the effectiveness of modeling and training agents for workforce optimization, our framework can be adapted to other similar service domains. One such example is of an organization responsible for delivering healthcare service to patients after they are discharged from hospital but need to cared for at their home. Facility in this example maps to the patient home locations, and service personnel maps to the healthcare provider staff i.e., physicians and nurses who travel to patient's home to assess and treat them. Similar to the example of service workforce management where a facility goes in distress and needs service, patient's health may deteriorate and need to be visited by the healthcare staff.

Another example is of an organization that is providing radiology staff to several radiology centers over a geographical area (e.g., city) to help them obtain medical scans for the incoming patients. The facility in this example maps to the radiology equipment at a hospital, and service personnel are the radiology staff that are trained to obtain medical scans. Note that unlike previous examples, the service requests are generated when a physician orders a medical scan of patients after examining them. However, the strategic management, positioning and dispatch of the radiology staff is necessary to achieve a balanced workforce cost and utilization while ensuring low patient wait time. 

\section{Conclusion and Future Work}
We presented an integrated simulation environment for workforce optimization and demonstrated that jointly training RL agents simultaneously addressing different aspects of the problem outperforms agents that are trained to optimize a single aspect (e.g., personnel dispatch). While the approach with jointly trained agents also outperforms manually designed heuristics, we envision that the provided simulation environment would facilitate RL research to bring further improvements in terms of scale (world size as well as number of facilities), thus leading to increased usage and value added by of RL in this space. Moving forward we intend to extend the framework to further include additional aspects of workforce environment such as personnel routing to optimize visits to multiple facilities, as well as, analyzing the tolerance of the jointly trained models to non-stationary variations in facility downtime frequency and travel durations. Other future extensions include accounting for hybrid scenarios, wherein personnel can provide on-site or remote service depending on the facility's need, which reflect certain service scenarios. With the growing trend of tackling large joint optimization spaces with RL, combined with the ever-increasing complexity of modern service organizations, we strongly believe that this work provides a meaningful step towards enabling the next-generation of AI for workforce management. \\

\noindent \textit{Disclaimer}: The concepts and information presented in this paper are based on research results that are not commercially available. Future commercial availability cannot be guaranteed.

\bibliography{rlwo}

\end{document}